%% file: paper.tex
\documentclass{llncs}
\usepackage{epstopdf}
\usepackage{graphicx}
\usepackage{epsf}

\begin{document}

\title{A Deep Learning Approach for Multimodal Deception Detection}

\author{Gangeshwar Krishnamurthy\inst{1} \and Navonil Majumder\inst{2} \and Soujanya Poria\inst{3} \and Erik Cambria\inst{4}}

\institute{A*STAR Artificial Intelligence Initiative (A*AI),\\
	Institute of High Performance Computing, Singapore\\
\email{gangeshwark@ihpc.a-star.edu.sg}
\and
Centro de Investigacin en \\
Computacin, IPN, Mexico\\
\email{navonil@sentic.net}
\and
	Temasek Laboratories,\\
	Nanyang Technological University, Singapore\\
\email{sporia@ntu.edu.sg}
\and
School of Computer Science and Engineering, \\
Nanyang Technological University, Singapore\\
\email{cambria@ntu.edu.sg}}

\maketitle

\begin{abstract}

Automatic deception detection is an important task that has gained momentum in computational linguistics due to its potential applications. In this paper, we propose a simple yet tough to beat multi-modal neural model for deception detection. By combining features from different modalities such as video, audio, and text along with Micro-Expression features, we show that detecting deception in real life videos can be more accurate. Experimental results on a dataset of real-life deception videos show that our model outperforms existing techniques for deception detection with an accuracy of 96.14\% and ROC-AUC of 0.9799.

\end{abstract}

\section{Introduction}
We face deceptive behavior in our day-to-day life. People lie to escape from a situation that seems unfavorable to them. As a consequence, some lies are innocuous but others may have severe ramifications in the society. Reports suggest that the ability of humans to detect deception without special aids is only 54\%\cite{bond2006accuracy}. A study by DePaulo et al. \cite{depaulo2003cues} found that deception without any particular motivation or intention exhibited almost no detectable cues of deception. However, cues were significantly more when lies were about transgressions.


With the rise in the number of criminal cases filed every year in the US\footnote{\url{http://www.uscourts.gov}}, it is ethically and morally important to accuse only the guilty defendant and free the innocent. Since the judgment for any case is mostly based on the hearings and evidence from the stakeholders (accused, witnesses, etc.), the judgment is most likely to go wrong if the stakeholders do not speak the truth. It is, hence, important to detect deceptive behavior accurately in order to upkeep the law and order.

Social media can be characterized as a virtual world where people interact with each other without the human feel and touch. It is easy to not reveal one's identity and/or pretend to be someone else on the social media. Cyberbullying is increasingly becoming a common problem amongst the teenagers nowadays \cite{teenage-cyberbullying}. These include spreading rumors about a person, threats, and sexual harassment. Cyberbullying adversely affects the victim and leads to a variety of emotional responses such as lowered self-esteem, increased suicidal thoughts, anger, and depression\cite{hinduja2014bullying}. Teenagers fall prey to these attacks due to their inability to comprehend the chicanery and pretentious behavior of the attacker.

Another area where deception detection is of paramount importance is with the increased number of false stories, a.k.a Fake News, on the Internet. Recent reports suggest that the outcome of the U.S. Presidential Elections is due to the rise of online fake news. Propagandists use arguments that, while sometimes convincing, are not necessarily valid. Social media, such as Facebook and Twitter, have become the propellers for this political propaganda. Countries around the world, such as France \cite{france-fakenews}, are employing methods that would prevent the spread of fake news during their elections. Though these measures might help, there is a pressing need for the computational linguistics community to devise efficient methods to fight Fake News given that humans are poor at detecting deception.


This paper is organized as follows. In section \ref{sec:related-work}, we will talk about the past work in deception detection; section \ref{sec:approach} describes our approach to solving deception detection. Section \ref{sec:experiment} explains our experimental setup. In section \ref{sec:results} and \ref{sec:drawbacks}, we discuss our results and drawbacks respectively. And finally, conclude with future work in section \ref{sec:conclusion}.

\section{Related Work}
\label{sec:related-work}

Past research in the detection of deception can be broadly classified as Verbal and Non-verbal. In verbal deception detection, the features are based on the linguistic characteristics, such as n-grams and sentence count statistics \cite{mihalcea2009linguistic}, of the statement by the subject under consideration. Use of more complex features such as psycholinguistic features \cite{pennebaker2001linguistic} based on the Linguistic Inquiry and Word Count (LIWC) lexicon, have also been explored by \cite{mihalcea2009linguistic} and shown that they are helpful in detecting deceptive behavior. Yancheva and Rudzicz studied the relation between the syntactic complexity of text and deceptive behavior \cite{P13-1093}.
1
In non-verbal deception detection, physiological measures were the main source of signals for detecting deceptive behavior. Polygraph tests measure physiological features such as heart rate, respiration rate, skin temperature of the subject under investigation. But these tests are not reliable and often misleading as indicated by \cite{vrij2000detecting,gannon2009risk} since judgment made by humans are often biased. Facial expressions and hand gestures were found to be very helpful in detecting deceptive nature. Ekman \cite{ekman2009telling} defined micro-expressions as short involuntary expressions, which could potentially indicate deceptive behavior. Caso et al. \cite{caso2006impact} identified particular hand gesture to be important to identify the act of deception. Cohen et al. \cite{cohen2010nonverbal} found that fewer iconic hand gestures were a sign of a deceptive narration.

Previous research was focused on detecting deceit behavior under constrained environments which may not be applicable in real life surroundings. Recently, the focus has been towards experiments in real life scenarios. Towards this, P{\'e}rez-Rosas et al. \cite{perez2015deception} introduced a new multi-modal deception dataset having real-life videos of courtroom trials. They demonstrated the use of features from different modalities and the importance of each modality in detecting deception. They also evaluated the performance of humans in deception detection and compared it with their machine learning models. Wu et al. \cite{larry2017deception} have developed methods that leverage multi-modal features for detecting detection. Their method heavily emphasizes on feature engineering along with manual cropping and annotating videos for feature extraction.

In this paper, we describe our attempt to use neural models that uses features from multiple modalities for detecting deception. We believe our work is the first attempt at using neural networks for deceit classification. We show that with the right features and simple models, we can detect deceptive nature in real life trial videos more accurately.

\section{Approach}
\label{sec:approach}

\subsection{Multi-Modal Feature Extraction}
The first stage is to extract unimodal features from each video. We extract textual, audio and visual features as described below.

\subsubsection{Visual Feature Extraction}
For extracting visual features from the videos, we use 3D-CNN \cite{ji20133d}. 3D-CNN has achieved state-of-the-art results in object classification on tridimensional data \cite{ji20133d}. 3D-CNN not only extracts features from each image frame, but also extracts spatiotemporal features \cite{tran2015learning} from the whole video which helps in identifying the facial expressions such as smile, fear, or stress.

The input to 3D-CNN is a video $v$ of dimension ($c, f, h, w$), where $c$ represents the number of channels and $f, h, w$ are the number of frames, height, and width of each frames respectively. A 3D convolutional filter, $f_l$ of dimension ($f_m, c, f_d, f_h, f_w$) is applied, where $f_m$ = number of feature maps, $c$ = number of channels, $f_d$ = number of frames (also called depth of the filter), $f_h$ = height of the filter, and $f_w$ = width of the filter. This filter, $f_l$, produces an output, $convout$ of dimension ($f_m, c, f-f_d+1, h-f_h+1, w-f_w+1$) after sliding across the input video, $v$. Max pooling is applied to $convout$ with window size being ($m_p, m_p, m_p$). Subsequently, this output is fed to a dense layer of size $d_f$ and softmax. The activations of this dense layer is used as the visual feature representation of the input video, $v$.

In our experiments, we consider only RBG channel images, hence $c = 3$. We use 32 feature maps and 3D filters of size, $f_d = f_h = f_w = 5$. Hence, the dimension of the filter, $f_l$, is $32 \times 3 \times 5 \times 5 \times 5$ . The window size, $m_p$, of the max pooling layer is 3.
Finally, we obtain a feature vector, $v_f$, of dimension 300 for an input video, $v$.

\subsubsection{Textual Features Extraction}
We use Convolutional Neural Networks (CNN) \cite{kim2014convolutional,kalchbrenner2014convolutional} to extract features from the transcript of a video, $v$. First, we use pre-trained Word2Vec \cite{mikolov2013distributed} model to extract the vector representations for every word in the transcript. These vectors are concatenated and fed as input vector to the CNN. We use a simple CNN with one convolutional layer and a max-pooling layer, to get our sentence representation. In our experiments, filters of size 3, 5 and 8 with 20 feature maps each is used. Window-size of 2 is employed for max-pooling over these feature maps. Subsequently, a full-connected layer with 300 neurons is used with rectified linear unit (ReLU) \cite{nair2010rectified} as the activation function. The activations of this full-connected layer is used as the textual feature representation of the input video, $v$. Finally, we obtain a feature vector, $t_f$, of dimension 300 for an input text (transcript), $t$.

\subsubsection{Audio Feature Extraction}
openSMILE \cite{Eyben2013opensmile} is an open-source toolkit used to extract high dimensional features from an audio file. In this work, we use openSMILE to extract features from the input audio. Before extracting the features, we make sure that there are no unnecessary signals in the audio that affects the quality of the extracted features. Hence, the background noise is removed from the audio and Z-standardization is used to perform voice normalization. To remove the background noise, we use SoX (Sound eXchange) \cite{sox} audio processing tool. The noiseless input audio is then fed to the openSMILE tool to extract high-dimensional features. These features are functions of low-level descriptor (LLD) contours. Specifically, we use the \emph{IS13-ComParE} openSMILE configuration to extract features which are of dimension $6373$ for every input audio, $a$.

After these features are extracted, a simple fully-connected neural network is trained to reduce the dimension to 300. Finally, we obtain a feature vector, $a_f$, of dimension 300 for an input audio, $a$.

\subsubsection{Micro-Expression Features}
Veronica et al. manually annotated facial expressions and use binary features derived from the ground truth annotations to predict deceptive behavior. Facial micro-expressions are also considered to play an important role in detecting deceptive behavior. The data provided by \cite{perez2015deception} contains 39 facial micro-expressions such as frowning, smiling, eyebrows raising, etc. These are binary features and taken as a feature vector, $m_p$ of dimension 39.

\begin{figure}[t!]
	\includegraphics[width=\linewidth]{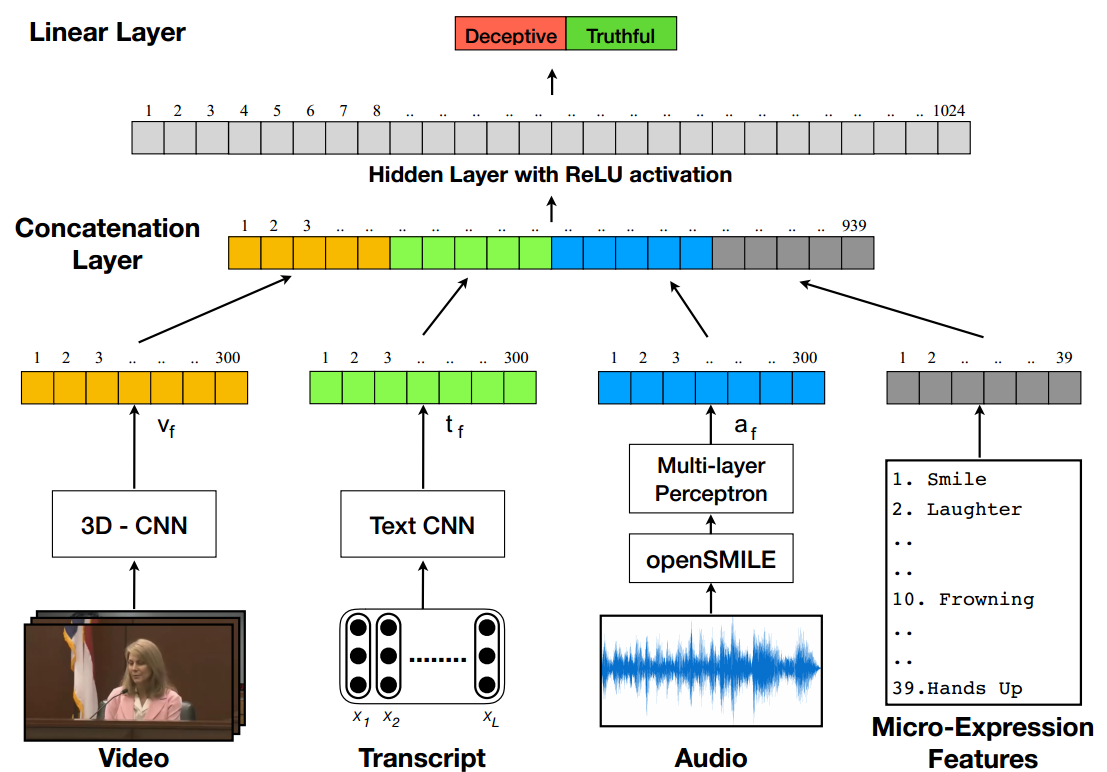}
	\caption{Architecture of model $MLP_C$}
	\label{fig:arch1}
\end{figure}

\begin{figure}[t!]
	\includegraphics[width=\linewidth]{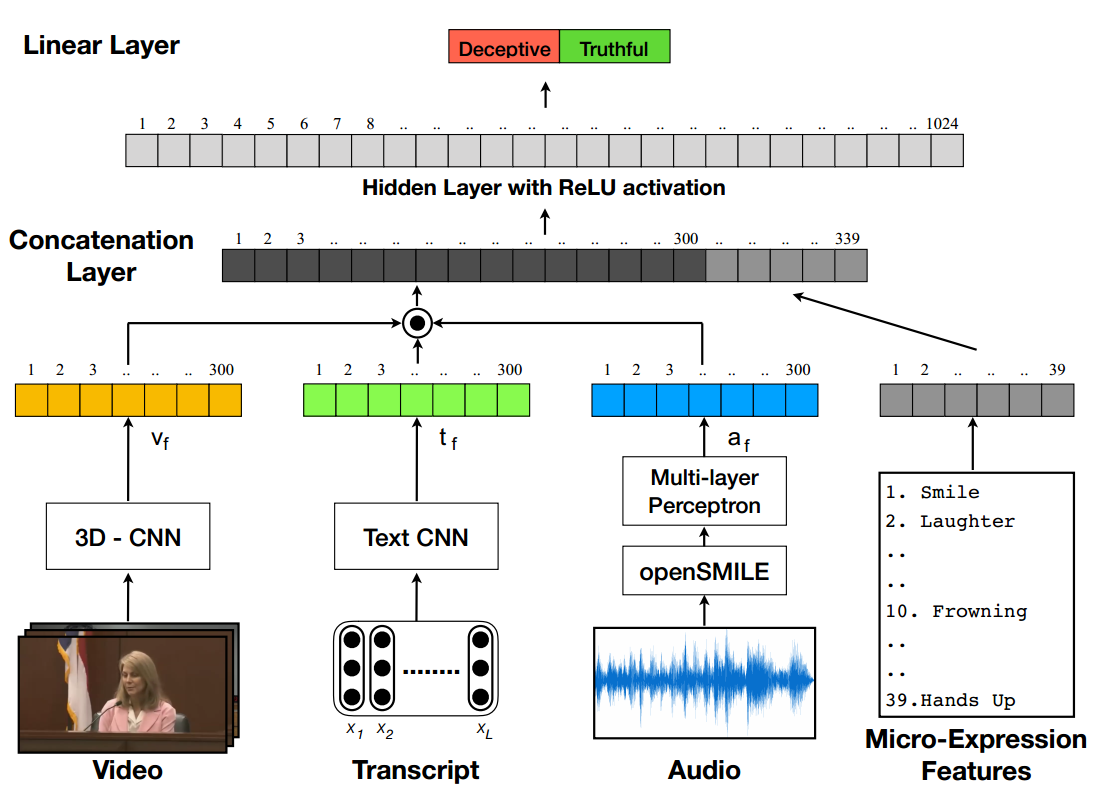}
	\caption{Architecture of model $MLP_{H+C}$}
	\label{fig:arch2}
\end{figure}

\subsection{Model Description}

\subsubsection{Multi-modal Model}
We use a simple Multi-Layer perceptron (MLP) (cite) with hidden layer of size 1024 followed by a linear output layer. We use the rectified linear unit (ReLU) activation function \cite{nair2010rectified} for non-linearity at the hidden layer. A dropout \cite{srivastava2014dropout} of keep probability, $p = 0.5$, is applied to the hidden layer for regularization.

Figure \ref{fig:arch1} and \ref{fig:arch2} shows the architecture of our models, $MLP_C$ and $MLP_{H+C}$.

\subsubsection{Unimodal Models}
We perform the evaluation on individual modalities and use the same architecture as multi-modal model. The only difference is that the input is either $v_f$, or $a_f$, or $t_f$, or $m_f$. Hence no data fusion is performed. We name the model as $MLP_U$.

\subsection{Data Fusion}
We fuse the features from individual modalities to map them into a joint space. To achieve this, we try different kinds of data fusion techniques:
\subsubsection{Concatenation}
In this method, the features from all the modalities are simply concatenated into a single feature vector. Thus, the extracted features, $t_f$, $a_f$, $v_f$ and $m_f$, are simply concatenated to form the representation: $z_f = [t_f; a_f; v_f; m_f]$ of dimension $d_{in} = 939$.

We call this model configuration as $MLP_{C}$ and is shown in the figure \ref{fig:arch1}.
\subsubsection{Hadamard + Concatenation}
In this method, the audio features, visual features and, textual features are fusion by using hadamard product. Then the Micro-Expression features are concatenated with the product. Thus, we have $z_f = [t_f \odot a_f \odot v_f; m_f]$ of dimension $d_{in} = 339$, where $(A \odot B)$ is element-wise multiplication between matrices $A$ and $B$ (also known as Hadamard product).

We call this model configurtion as $MLP_{H+C}$ and is shown in the figure \ref{fig:arch2}.

\subsection{Loss}
The training objective is to minimise the cross-entropy between the model's output and the true labels. We trained the models with back-propagation using Stochastic Gradient Descent optimizer. The loss function is:

\begin{equation}
	J = \frac{-1}{N}{\sum_{i=1}^{N}}{\sum_{j=1}^{C}}{y_{i,j} \ log_2 (\hat{y}_{i,j} )} 
\end{equation}
Here, $N$ is the number of samples, $C$ is the number of categories (in our case, $C = 2$). $y_{i}$ is the one-hot vector ground truth of $i^{th}$ sample and $\hat{y}_{i,j} $ is its predicted probability of belonging to class $j$.

\section{Experiment}
\label{sec:experiment}
\subsection{Data}
For evaluating our deception detection model, we use a real-life deception detection dataset by \cite{perez2015deception}. This dataset contains 121 video clips of courtroom trials. Out of these 121 videos, 61 of them are of deceptive nature while the remaining 60 are of truthful nature. 


The dataset contains multiple videos from one subject. In order to avoid bleeding of personalities between train and test set, we perform a 10-fold cross validation with subjects instead of videos as suggested by Wu \emph{et al.} \cite{larry2017deception}. This ensures that videos of the same subjects are not in both training and test set.

\subsection{Baselines}
Wu \emph{et al.} \cite{larry2017deception} have made use of various classifiers such as Logistic Regression (LR), Linear SVM (L-SVM), Kernel SVM (K-SVM). They report the AUC-ROC values obtained by the classifiers for different combination of modalities. We compare the AUC obtained by our models against only Linear SVM (L-SVM) and Logistic Regression (LR).

P{\'e}rez-Rosas \emph{et al.} \cite{perez2015deception} use Decision Trees (DT) and Random Forest (RF) as their classifiers. We compare the accuracies of our models against DT and RF.

\begin{table}
	\caption{Comparision of AUC}
	\label{tab:table1}
	\begin{center}
		\begin{tabular}{|l|c|c||c|c|c|}
			\hline
			\bf Features & L-SVM \cite{larry2017deception} & LR \cite{larry2017deception} & \bf $MLP_U$ & $MLP_C$ & $MLP_{H+C}$ \\
			\hline
            \bf Random  & - & - & 0.4577 & 0.4788 & 0.4989 \\ \hline
			\bf Audio  & \bf 0.7694 & 0.6683 & 0.5231 & - & - \\
			\bf Visual & 0.7731 & 0.6425 & \bf 0.9596 & - & - \\
			\bf Textual (Static)& 0.6457 & 0.5643 & \bf 0.8231 & - & -\\
			\bf Textual (Non-static)& - & - & \bf 0.9455 & - & -\\
			\bf Micro-Expression & 0.7964 & \bf 0.8275 & 0.7512 & - & -\\
			\hline \hline
			\bf All Features (Static)  & 0.9065 & 0.9221 & - & 0.9033 & \bf 0.9348 \\
			\bf All Features (Non-static) & - & - & - & 0.9538 & \bf 0.9799 \\
			\hline
		\end{tabular}
	\end{center}
\end{table}

\section{Results}
\label{sec:results}
Tables \ref{tab:table1} \& \ref{tab:table2} presents the performances of $MLP$ and its variants along with the state-of-the-art models. During feature extraction from text, we train the TextCNN model with two different settings: one, by keeping the word vector representation static; two, by optimizing the vector along with the training (non-static). In our results, we also show the performance of the model with these two textual features separately. Additionally, we also mention the results we got from our models for feature vectors initialized with random numbers.

Table \ref{tab:table1} shows that our model, $MLP_{H+C}$, obtains an AUC of 0.9799 while outperforming all other competitive baselines with a huge margin.

\begin{table}
	\caption{Comparing accuracies of our model with baselines}
	\label{tab:table2}
	\begin{center}
		\begin{tabular}{|l|c|c||c|c|c|}
			\hline
			\bf Features & DT \cite{perez2015deception} & RF \cite{perez2015deception} & \bf $MLP_U$ & $MLP_C$ & $MLP_{H+C}$ \\
			\hline
            \bf Random  & - & - & 43.90\% & 45.32\% & 48.51\% \\ \hline
			\bf Audio  & - & - & \bf 52.38\% & - & - \\
			\bf Visual & - & - & \bf 93.08\% & - & - \\
			\bf Textual (Static) & 60.33\% & 50.41\% & \bf 80.16\% & - & - \\
			\bf Textual (Non-static) & - &  - & \bf 90.24\% & - & - \\
			\bf Micro-Expression & 68.59\% & 73.55\% & \bf 76.19\% & - & - \\
			\hline \hline
			\bf All Features (Static) & 75.20\% & 50.41\% & - & 90.49\% & \bf 90.99\% \\
			\bf All Features (Non-static) & - & - & - & 95.24\% & \bf 96.14\% \\
			\hline
		\end{tabular}
	\end{center}
\end{table}

Table \ref{tab:table2} compares the performance our models with Decision Tree and Linear Regression models \cite{perez2015deception}. Our model, $MLP_{H+C}$, again outperforms other baselines by achieving an accuracy of 96.14\%. We can also infer that visual and textual features play a major role in the performance of our models; followed by Micro-Expressions and audio. This conforms with the findings by \cite{perez2015deception} that facial display features and unigrams contribute the most to detecting deception. 

As we can see that, our approach outperforms the baselines by a huge margin. Our neural models simple and straightforward, hence the results show that right feature extraction methods can help in unveiling significant signals that are useful for detecting deceptive nature.

\section{Drawbacks}
\label{sec:drawbacks}
Though our models outperform the previous state-of-the-art models, we still acknowledge the drawbacks of our approach as follows:
\begin{itemize}
\item Our models still rely on a small dataset with only 121 videos. Due to this, our models are prone to over-fitting if not carefully trained with proper regularization.
\item Also, due to the limited scenarios in the dataset, the model may not show the same performance for out-of-domain scenarios.
\end{itemize}


\section{Conclusions and Future Work}
\label{sec:conclusion}
In this paper, we presented a system to detect deceptive nature from videos. Surprisingly, our model performed well even with only 121 videos provided in the dataset, which is generally not a feasible number of data points for neural models. As a result, we conclude that there exists a certain pattern in the videos that provide highly important signals for such precise classification. We performed various other evaluations not presented in this paper, to confirm the performance of our model. From these experiments, we observed that visual and textual features predominantly contributed to accurate predictions followed by Micro-Expression features. Empirically, we observed that our model $MLP_{H+C}$ converged faster in comparison with $MLP_C$.

While our system performs well on the dataset by \cite{perez2015deception}, we can not claim the same performance of our model for larger datasets covering a larger number of environments into consideration. Hence, creating a large multi-modal dataset with a large number of subjects under various environmental condition is part of our future work. This would pave a way to build more robust and efficient learning systems for deception detection. Another interesting path to explore is detecting deception under social dyadic conversational setting.




\bibliographystyle{splncs}
\bibliography{paper}
\end{document}